\documentclass[lettersize,journal]{IEEEtran}
\let\theoremstyle\relax

\usepackage{amsthm}
\theoremstyle{plain}
\newtheorem{theorem}{Theorem}[section]
\newtheorem{lemma}[theorem]{Lemma}

\theoremstyle{definition}
\newtheorem{definition}[theorem]{Definition}
\theoremstyle{remark}

\usepackage{cite}
\usepackage{amsmath,amssymb,amsfonts}
\usepackage{graphicx}
\usepackage{subfigure}
\usepackage{color}

\usepackage{multirow}
\usepackage[ruled,vlined]{algorithm2e}
\usepackage{cuted}
\usepackage[mathscr]{euscript}

\begin{document}

\title{A Non-Classical Parameterization for Density Estimation Using Sample Moments}

\author{Guangyu Wu,~\IEEEmembership{Student Member,~IEEE}, and Anders Lindquist, \IEEEmembership{Life Fellow, IEEE}
        % <-this % stops a space
\thanks{Guangyu Wu is with Department of Automation, Shanghai Jiao Tong University, Shanghai, China. (e-mail: chinarustin@sjtu.edu.cn).}
\thanks{Anders Lindquist is with Department of Automation and School of Mathematical Sciences, Shanghai Jiao Tong University, Shanghai, China. (e-mail: alq@kth.se).}}

% The paper headers
\markboth{Journal of \LaTeX\ Class Files,~Vol. , No. , ~ }%
{Shell \MakeLowercase{\textit{et al.}}: A Sample Article Using IEEEtran.cls for IEEE Journals}

% \IEEEpubid{0000--0000/00\$00.00~\copyright~2021 IEEE}
% Remember, if you use this you must call \IEEEpubidadjcol in the second
% column for its text to clear the IEEEpubid mark.

\maketitle

\begin{abstract}
Probability density estimation is a core problem of statistics and signal processing. Moment methods are an important means of density estimation, but they are generally strongly dependent on the choice of feasible functions, which severely affects the performance. In this paper, we propose a non-classical parametrization for density estimation using sample moments, which does not require the choice of such functions. The parametrization is induced by the squared Hellinger distance, and the solution of it, which is proved to exist and be unique subject to a simple prior that does not depend on data, and can be obtained by convex optimization. Statistical properties of the density estimator, together with an asymptotic error upper bound are proposed for the estimator by power moments. Applications of the proposed density estimator in signal processing tasks are given. Simulation results validate the performance of the estimator by a comparison to several prevailing methods. To the best of our knowledge, the proposed estimator is the first one in the literature for which the power moments up to an arbitrary even order exactly match the sample moments, while the true density is not assumed to fall within specific function classes.
\end{abstract}


\section{Introduction}

\IEEEPARstart{D}{ensity} estimation is an important problem of statistics and signal processing, which also lies in the core of numerous machine learning tasks, e.g. clustering and generative modeling. It can be formulated as follows. Given a set of independent and identically distributed (i.i.d.) samples from an unknown true distribution, find a density estimate that best describes the true one.

Since no prior information about the density function is given other than the data samples, it has been considered infeasible to treat the density estimation problem unless assuming the densities to fall within specific classes of functions, which we call a parametrization of the density. The mixture models, such as Parzen windows \cite{parzen1962estimation, silverman2018density} or mixtures of Gaussians or other basis functions \cite{mclachlan1988mixture, bunea2007sparse} are parameterized as mixtures of kernel functions, of which the type and the bandwidth need to be chosen carefully. However the performance of nonparametric algorithms is quite limited when the sample size is small.

On the other hand, power moments have been used to characterize the data samples. Methods matching the moments of the estimators to those of the data have been proposed in several papers \cite{barndorff2014information, dudik2004performance, altun2006unifying}. However, these density estimators employ exponential family models, and the feasible density classes of these methods are very limited. The moment matching method for nonparametric mixture models proposed in \cite{song2008tailoring} brings flexibility to the conventional moment methods, but a good knowledge of the function class is still required. Moreover, the existence of solution has not been proved in the previous papers. Neither are the statistical properties and error upper bounds proved, which severely lowers the value of those algorithms in application.

In conclusion, how to parameterize the density estimates given the samples is one of most significant problems in density estimation. In a long series of contributions, the parametrization has been separated into several small tasks. For example, mode estimation is about estimating the modes of a distribution, e.g. \cite{parzen1962estimation, chernoff1964estimation, eddy1980optimum, silverman2018density, cheng1995mean, abraham2004asymptotic, dasgupta2014optimal, genovese2012minimax, jiang2017modal}, with modes viewed as the central tendencies of a distribution. Class probability estimation involves estimating the probability distribution over a set of classes for a given input \cite{rigollet2007generalization}, etc. These results made significant contributions to the parametrization problem. However, since all of the tasks will bring individual biases to the parametrization, a parametrization of densities with minimum requirement of individual prior constraints, for example, the modes of the distribution and the function classes, is of great interest. 

A parametrization for spectral density estimation using sample moments by Kullback-Leibler distance has been proposed in \cite{georgiou2003kullback}. It only requires a prior spectral density independent of the samples. However in this problem, the number of data samples is limited. It makes the Kullback-Leibler divergence no longer the most satisfactory criterion to estimate the probability density functions, since especially  it depends  sensitively on events that are very rare in the reference distribution, which may induce sharp peaks in the density estimates. We naturally consider other metrics for density estimation using sample moments.

This paper treats the problem of density estimation by sample moments. The density estimation problem is formulated as a truncated Hamburger moment problem, and a solution to the proposed problem is proved to exist. The problem is then formulated in a matrix equation form and the squared Hellinger distance are used to form a convex optimization problem, and a parametrization of a rational form is proved to be the unique solution of it by proving the map from parameters of the parametrization to the sample moments being homeomorphic, which also makes it possible to apply gradient-based algorithms to treat the convex optimization problem. Then we prove the statistical properties of the proposed estimator. An asymptotic error upper bound of the estimator is also derived. Then we propose to use the density estimator in two signal processing tasks, including the observation noise estimation for filtering, and the recursive Bayesian estimator with non-Gaussian density functions. We also explain why the proposed estimator is indispensable in these tasks. Last but not the least, the simulation results of density estimation on mixtures of Gaussians, Laplacians and Gumbels are given, which validate the proposed density estimator. We emphasize that our density estimator can treat multi-modal densities without estimation/prior knowledge of modes or feasible classes.

\section{Problem formulation}

In this section, we will formulate our problem of density estimation using sample moments. We assume the density estimate to be supported on $\mathbb{R}$. Then the problem falls within the Hamburger moment problem \cite{schmudgen2017moment, bertsimas2005optimal}. 

In the conventional Hamburger moment problem, one investigates whether a sequence is a feasible moment sequence. A sequence $\mu = (\mu_0, \mu_1, \dots, \mu_\nu)$
is a feasible $\nu$-sequence, if there exists a random variable $X$ with a probability density function $p(x)$ supported on $\mathbb{R}$, of which the power moments are given by
$$
\mu_{k} = \mathbb{E}\{X^{k} \} =\int_\mathbb{R}x^k p(x)dx, \quad k=0,1,\dots,\nu.
$$
Then any such random variable $X$ is said to have a $\mu$-feasible distribution and is denoted as $X \sim\sigma$. 

However, in density estimation, we need an estimate of the probability density $p(x)$, a problem which may have infinitely many solutions. In this paper, we shall deal with a moment estimation problem to distinguish it from the conventional Hamburger moment problem. We should always remember that order $\nu$ moment estimation problem is ill-posed. Only if proper constraints are given, an analytic solution to the Hamburger moment problem can be obtained. 
Moreover, rather than the true moment sequence, we treat the Hamburger moment problem with a sample power moment sequence.

\begin{definition}[Order $2n$ moment density estimation problem]
Given a sequence $\mu = \left( \mu_{0}, \cdots, \mu_{2n} \right)$ with
\begin{equation}
  \mu_{k} = \frac{1}{m}\sum_{j=1}^{m} X^{k}_{j}, \quad k=0, \ldots, 2n, 
\label{MomentCal}
\end{equation}
where $X_1, X_2,\dots,X_m$ are independent and identically distributed samples. $\mu$ is the sample moment sequence. The estimation problem is then to find a density estimate $p(x)$ corresponding to a random variable $X \sim \mu$.
\label{d2}
\end{definition}

Thus density estimation using the truncated moment sequence obtained from the samples has been formulated as a Hamburger moment problem. Before treating this problem, we first need to prove the existence of solutions.

\section{Existence of solutions}

Since we are using sample moments, which due to sampling errors differ from the true population moments of the density function to be estimated, we need to prove that there exists a solution to Definition \ref{MomentCal}. To this end, we review some facts about the solvability of the power moment problem. 

\begin{theorem}[Solution of the Hamburger Moment Problem \cite{schmudgen2017moment}] Denote the nonnegative integers as $\mathbb{N}_{0}$ and the positive Radon measures on the real numbers as $M_{+}(\mathbb{R})$. For a real sequence $s=\left(s_{n}\right)_{n \in \mathbb{N}_{0}}$ the following are equivalent:

\noindent (i) $s$ is a Hamburger moment sequence, that is, there is a Radon measure $\eta \in M_{+}(\mathbb{R})$ such that $x^{n} \in \mathcal{L}^{1}(\mathbb{R}, \eta)$ and
$$
s_{n}=\int_{\mathbb{R}} x^{n} d \eta(x) \text { for } n \in \mathbb{N}_{0}
$$

\noindent (ii) The sequence $s$ is positive semidefinite.

\noindent (iii) All Hankel matrices
\begin{equation*}
   H_{n}(s)=\left[\begin{array}{cccc}
    s_{0} & s_{1} & \ldots & s_{n} \\
    s_{1} & s_{2} & \ldots &s_{n+1} \\
    \vdots & \vdots & \ddots & \vdots \\
    s_{n} & s_{n+1} & \ldots & s_{2n}
    \end{array}\right], \quad n \in \mathbb{N}_{0}
\label{Hankel}
\end{equation*}
are positive semidefinite.
% i.e., $\sum_{j=0}^\infty\sum_{k=0}^\infty a_j s_{j+k}a_k \geq 0$ for all sequences $a$. 
\label{t31}
\end{theorem}

Next we shall prove that the truncated Hamburger moment problem in Definition \ref{d2} is solvable.

\begin{theorem}
The problem in Definition \ref{MomentCal} for $\mu$ with the moments given by (\ref{d2}) is solvable, if and only if $X_1, X_2,\dots,X_m$ are not all equal. Moreover, the sequence $\mu$ is positive definite.
\label{t32}
\end{theorem}

\begin{proof} We note that the empirical distribution function
$$
    \eta \left( x \right) = \frac{1}{m}\sum_{i=0}^{m}{\mathbb{I}}_{[{X}_{i},+\infty)} (x),
$$
where $\mathbb{I}$ is the indicator function, is a Radon measure. Then, by Theorem \ref{t31}, the sample moment sequence ${\mu} = \left ( {\mu}_{0}, {\mu}_{1}, \cdots {\mu}_{2n} \right )$ is a positive semidefinite sequence (because the full sample moment sequence is positive semidefinite). 
We note that a positive semidefinite sequence $\mu$ is positive definite if and only if $X_1, X_2,\dots,X_m$ are not all equal, which is an event of probability $1 - \int_{\mathbb{R}}\left(p(x)\right)^{m}dx$. Then by Corollary 9.2 in \cite{schmudgen2017moment}, we have that the truncated Hamburger moment problem for $\mu$ is solvable given that $X_1, X_2,\dots,X_m$ are not all equal.
\end{proof}

\section{An analytic solution by squared Hellinger distance}

In the previous section, solutions to the order $2n$ moment estimation problem is proved to exist (Theorem \ref{t32}). However, the existence of solutions doesn't necessarily guarantee analytic solutions to the moment problem. In this section, we will propose a method to obtain analytic solutions to this problem. In \cite{georgiou2003kullback}, the constraints on the sample moments were the positive definiteness of a Toeplitz matrix, Pick matrix or a similar object. In this paper, the appropriate Hankel matrix needs to be positive definite. Therefore, we write the Hamburger moment problem in a Hankel matrix form following some lines of thoughts in \cite{georgiou2003kullback, wu2023non}.

The power moment conditions up to order $2n$ read
\begin{displaymath}
\mu_{k} =\int_\mathbb{R}x^kp(x)dx, \quad k=0,1,\dots,2n.
\end{displaymath}
Then we can write them as a matrix equation
\begin{equation}
   \int_\mathbb{R} F(x)p(x) F^{T}(x)dx=\mathscr{M}_{2n} ,
\label{IntegG}
\end{equation}
where 
$$
F(x)=\left[\begin{array}{c}
1 \\
x \\
\vdots \\
x^{n-1} \\
x^{n}
\end{array}\right]
$$
and
$$
    \mathscr{M}_{2n}=\left[\begin{array}{cccc}
    \mu_{0} & \mu_{1} & \ldots & \mu_{n} \\
    \mu_{1} & \mu_{2} & \ldots & \mu_{n+1} \\
    \vdots & \vdots & \ddots & \vdots \\
    \mu_{n} & \mu_{n+1} & \ldots & \mu_{2n}
    \end{array}\right]
$$
with the power moments $\mu_{k}, k = 0, \cdots, 2n$,  calculated by (\ref{MomentCal}). $\mathscr{M}_{2n}$ is the Hankel matrix. By doing this, we give a formulation of Definition \ref{d2}.

We denote by $\mathcal{P}$ be space of all probability density functions supported on $\mathbb{R}$. Let $\mathcal{P}_{2n}$ be the subspace of all $p\in\mathcal{P}$ of which at least $2n$ orders of power moments (in addition to $\mu_0$, which is 1) exist and are finite. By Theorem \ref{t32}, the set of all feasible $p\in\mathcal{P}$ which satisfy (\ref{IntegG}) is nonempty and that $\mathscr{M}_{2n} \succ 0$. Moreover, $\mathscr{M}_{2n}$ falls within the range of a linear operator
\begin{equation}
    \Xi: p \mapsto \mathscr{M}_{2n}=\int_\mathbb{R} F(x) p(x) F^{T}(x)dx ,
\label{Map}
\end{equation}
defined on $\mathcal{P}_{2n}$.  Since $\mathcal{P}_{2n}$ is convex,  $\operatorname{range}(\Xi)=\Xi\mathcal{P}_{2n}$ is also convex.

In the literature, the Kullback-Leibler (KL) distance is a commonly used measure of the difference between probability density functions \cite{hall1987kullback, li1999mixture, wu2023non}. However it doesn't satisfy the symmetric condition for being a metric. Moreover, the Kullback–Leibler divergence depends especially sensitively on events that are very rare in the reference distribution. Always formulated as minimizing the distance between a prior density and a proposal density \cite{vapnik1999nature, georgiou2003kullback}, to use KL divergence as the bona-fide distance measure for density estimation doesn't always yield satisfactory estimates. 

Let $r$ be an arbitrary probability density in $\mathcal{P}$. In this paper, following some lines of thoughts of \cite{wu2023non}, we propose to use the squared Hellinger distance instead of the KL divergence, which is written as

\begin{equation}
   \mathbb{H}^{2}(r, p) = \int_{\mathbb{R}} \left( \sqrt{r(x)} - \sqrt{p(x)} \right)^{2} dx
\label{Hellinger}
\end{equation}
to consider the distance
between $r$ and $p$. There are several advantages to use the squared Hellinger distance. First it is jointly convex, and is a real distance metric. Second, it penalizes the estimation error in the sense of L2 norm. It may ameliorate the sharp peaks in the estimates, which is very common when the KL divergence is chosen as the distance measure. We emphasize that the choice of distance metric is not arbitrary. We observe that in the widely used measures, only the KL and the Hellinger distances yield an analytic parametrization for the order $2n$ moment density estimation problem.

Hellinger distance is also a widely used metric. However in the previous results, density estimation by Hellinger distance always needs a prescribed model, and the estimation is performed by estimating the parameters of the model \cite{cutler1996minimum, lu2003minimum}. In this section, we introduce a parametrization of $p\in\mathcal{P}_{2n}$,  which is induced by the squared Hellinger distance, but without any other estimation or prior knowledge of the modes and feasible density classes.

\begin{theorem}
Given $\Xi$ which is defined by (\ref{Map}), denote
\begin{equation*}
\label{Lplus}
\mathcal{R}_{+}:=\left\{\Omega \in \operatorname{range}(\Xi) \mid F\left(x\right)^{T} \Omega F\left(x\right)>0, x \in \mathbb{R} \right\}.
\end{equation*}
Minimizing (\ref{Hellinger})
subject to $\Xi(p)=\mathscr{M}_{2n}$, i.e., subject to (\ref{IntegG}), yields a unique $p \in \mathcal{P}_{2n}$ which has the form 
\begin{equation}
    \hat{p}=\frac{r}{(1 + F^{T} \hat{\Omega} F)^{2}},
\label{MiniForm}
\end{equation}
for any choice of $r \in \mathcal{P}$ and $\mathscr{M}_{2n} \succ 0$. Moreover, minimizing the functional
\begin{equation}
    \mathbb{J}_{r}(\Omega):=\operatorname{tr}(\Omega \mathscr{M}_{2n}) + \int_{\mathbb{R}} \frac{r}{1 + F^{T} {\Omega} F}dx
\label{CostFunc}
\end{equation}
over all $\Omega \in \mathcal{R}_{+}$ yields a unique $\hat{\Omega}$.
\label{Theorem41}
\end{theorem}

\begin{proof}
We write the dual functional
$$
L(p, \Omega)=\mathbb{H}^{2}(r,  p) + \operatorname{tr}(\Omega(\Xi(p)-\mathscr{M}_{2n})),
$$
where $\Omega \in\operatorname{range}(\Xi)$ is the Lagrange multiplier, then it amounts to maximizing the dual functional
\begin{equation}
\label{infL}
\Omega \mapsto \inf _{p \in \mathcal{P}_{2n}} L(p, \Omega).
\end{equation}
It is clear that $p\mapsto L(p, \Omega)$ is a strictly convex map. Therefore, to obtain the maximum of the right member of (\ref{infL}), a $p\in\mathcal{P}_{2n}$ satisfying $\delta L(p, \Omega ; \delta p) =0$ for all $\delta p$ needs to be obtained. This will further constrain the choice of $\Omega$. By denoting
\begin{equation}
\label{q}
\omega(x):=F(x)^T\Omega F(x) + 1,
\end{equation}
the dual functional can be written as
$$
\begin{aligned}
& L(p, \Omega)\\
= & \int_\mathbb{R} \left( \sqrt{r(x)} - \sqrt{p(x)} \right)^{2}dx + \int_\mathbb{R}\left(\omega(x)-1\right)p(x)dx\\
- & \operatorname{tr}(\Omega \mathscr{M}_{2n}).
\end{aligned}
$$
The directional derivative reads
$$
\delta L(p, \Omega ; \delta p)=\int_\mathbb{R} \delta p(x)\left(\omega(x) - 1 + 1 - \frac{\sqrt{r(x)}}{\sqrt{p(x)}}\right)dx,
$$
For a $p$ to be a minimum point, all variations $\delta p$ need to be zero, which is achieved only if

$$
    \omega(x) = \frac{\sqrt{r(x)}}{\sqrt{p(x)}}, \quad \text{i.e.}, \quad p(x) = \frac{r(x)}{\omega^{2}(x)}
$$
for all $x\in\mathbb{R}$. 
\end{proof}

Since $r(x)$ and $p(x)$ are supported on $\mathbb{R}$, which are both strictly positive, we have that $\omega(x)$ is also strictly positive. By (\ref{IntegG}) and (\ref{q}), we shall further constrain $\Omega\in\mathcal{R}_+$.

\begin{lemma}
    $\Omega \in \mathcal{R}_+$ only if $\omega(x)>0$.
\end{lemma}

\begin{proof}
Given $\Omega \in \mathcal{R}_+$, $\Omega$ can be written as
$$
   \int_\mathbb{R} F(x)\psi(x) F^{T}(x)dx=\Omega,
$$
where $\psi \in \mathcal{P}_{2n}$. Therefore we have
$$
   F^{T}\int_\mathbb{R} F\psi F^{T}dx F
   = F^{T}\Omega F = \omega - 1.
$$

Since $\omega(x)$ is a scalar, it can be written as
$$
\begin{aligned}
    \omega(x) & = \operatorname{tr}\left( F^{T}\Omega F \right) + 1\\
    & = \operatorname{tr}\left( F^{T}\int_\mathbb{R} F\psi F^{T}dx\cdot F \right) + 1\\
    & = \operatorname{tr}\left( F^{T}F\int_\mathbb{R} F\psi F^{T}dx \right) + 1\\
    & = F^{T}F\operatorname{tr}\left( \int_\mathbb{R} F\psi F^{T}dx \right) + 1\\
    & = F^{T}F \operatorname{tr}\left( \int_\mathbb{R} \sum_{i = 0}^{n} x^{2i}\psi(x)dx \right) + 1
\end{aligned}
$$
where $F^{T}F$ is a scalar. Since $x^{2i}, \psi$ and $F^{T}F$ are all positive, we have $\omega(x) > 0$.
\end{proof}

Then the dual functional reads
\begin{displaymath}
L\left(\frac{r}{q},\Omega\right)=-\mathbb{J}_{r}(\Omega) + \int_{\mathbb{R}}r(x) dx,
\end{displaymath}
where $\mathbb{J}_{r}$ is given by (\ref{CostFunc}). The dual problem is now to minimize $\mathbb{J}_{r}(\Omega)$ over $\mathcal{R}_+$. To complete the proof we still need to prove the following theorem.

\begin{theorem}
There exists a unique minimum $\hat{\Omega} \in\mathcal{R}_{+}$ for the functional $\mathbb{J}_{r}(\Omega)$ satisfying
$$
\Xi\left(\frac{r}{(1 + F^{T} \hat{\Omega} F)^{2}}\right)=\mathscr{M}_{2n}.
$$
\label{t43}
\end{theorem}
By this theorem, to be proved below,
$$
\hat{p}=\frac{r}{\hat{q}^{2}}
$$
where $\hat{q}=1 + F^{T}\hat{\Omega} F$ belongs to $\mathcal{P}_{2n}$. Moreover, $\mathcal{P}_{2n}$ is a stationary point of the strictly convex dual functional $p \mapsto L(p, \hat{\Omega})$. Therefore, we have
$$
L(\hat{p}, \hat{\Omega}) \leq L(p, \hat{\Omega}), \quad \forall p \in \mathcal{P}_{2n}.
$$
Since $\Xi(\hat{p})=\mathscr{M}_{2n}$,
\begin{equation}
\mathbb{H}^{2}(r, \hat{p}) \leq \mathbb{H}^{2}(r, p)
\label{Hthetarho}
\end{equation}
for all $p\in \mathcal{P}_{2n}$ satisfying the constraint (\ref{IntegG}). The necessary and sufficient condition of the equality in (\ref{Hthetarho}) is $p=\hat{p}$. Theorem \ref{Theorem41} is proved.

Next, we shall finish the proof of Theorem \ref{t43}. We first investigate the dual problem of minimizing $\mathbb{J}_{r}(\Omega)$ over $\mathcal{R}_{+}$.

\begin{lemma}
Define the map $\zeta: \mathcal{R}_{+} \rightarrow \mathcal{X}_{+}$ between $\mathcal{R}_{+}$ and $\mathcal{X}_{+}:=\{\mathscr{M}_{2n} \in \operatorname{range}(\Xi)\mid \mathscr{M}_{2n} \succ 0\}$ as
$$
\zeta: \; \Omega \mapsto \int_\mathbb{R} F(x) \frac{r(x)}{\omega^{2}(x)} F(x)^{T}dx
\label{omega}
$$
where $\omega(x)$ is defined by (\ref{q}). An $\Omega$ is a stationary point of $\mathbb{J}_{r}(\Omega)$ only if
\begin{equation}
    \zeta(\Omega)=\mathscr{M}_{2n}.
    \label{Omega}
\end{equation}
\end{lemma}

\begin{proof}
By (\ref{CostFunc}) and (\ref{q}) we have
\begin{displaymath}
    \mathbb{J}_{r}(\Omega):=\operatorname{tr}(\Omega \mathscr{M}_{2n}) + \int_{\mathbb{R}} \frac{r}{1 + F^{T} {\Omega} F}dx
\end{displaymath}
and therefore by
\begin{displaymath}
\delta q(\Omega;\delta\Omega)=F^T\delta\Omega F = \operatorname{tr}\{\delta\Omega FF^T\},
\end{displaymath}
the directional derivative reads
\begin{equation}
\begin{aligned}
& \delta \mathbb{J}_{r}(\Omega ; \delta \Omega)\\
= & \operatorname{tr}\left(\delta \Omega\left[\mathscr{M}_{2n}-\int_\mathbb{R} F(x) \frac{r(x)}{\omega^{2}(x)} F(x)^{T}dx\right]\right).
\end{aligned}
\label{FirstOrder}
\end{equation}

The necessary and sufficient condition for (\ref{FirstOrder}) being zero over all $\delta \Omega \in\operatorname{range}(\Xi)$ is (\ref{Omega}). The proof is complete. 
\end{proof}

We also need to establish the injectivity and surjectivity of the map $\zeta: \mathcal{R}_{+} \mapsto \mathcal{X}_{+}$ for the proof of Theorem \ref{t43}. By doing this, we will prove that there exists a unique solution of (\ref{Omega}), and then equivalently the dual functional $\mathbb{J}_{r}$ has a unique minimum. We begin with injectivity.

\begin{lemma}
    The dual functional $\mathbb{J}_{r}(\Omega)$ is strictly convex.
\label{l46}
\end{lemma}

\begin{proof}
    It requires to prove $\delta^{2} \mathbb{J}_{r} > 0$ where
\begin{equation}
\delta^{2} \mathbb{J}_{r}(\Omega; \delta \Omega)=\int_\mathbb{R} \frac{2r(x)}{\omega(x)^{3}}\left(F(x)^{T} \delta\Omega F(x)\right)^{2}dx
\label{SecondDeriv}
\end{equation}
By (\ref{SecondDeriv}), we have $\delta^{2} \mathbb{J}_{r} \geq 0$. Now we still need to prove that
$$
    \delta^{2} \mathbb{J}_{r} > 0, \quad \text{for all} \quad \delta \Omega \neq \mathbf{0}.
$$
By Lemma 3.6 in \cite{wu2023non}, it is proved that $\Omega \mapsto F^{T} \Omega F$ is injective. By replacing $\Omega$ with $\delta\Omega$, this lemma is proved.
\end{proof}

It follows from Lemma \ref{l46} that there is only one stationary point satisfying (\ref{Omega}), i.e., the map $\zeta: \mathcal{R}_{+} \rightarrow \mathcal{X}_{+}$ is injective. 

Next, we shall prove the surjectivity of $\zeta: \mathcal{R}_{+} \rightarrow \mathcal{X}_{+}$. A similar proof can be found in \cite{wu2023non} and we omit it here. The injectivity and surjectivity proofs complete the proof of Theorem \ref{t43}. Therefore $\zeta: \mathcal{R}_{+} \rightarrow \mathcal{X}_{+}$ is a proper and injective continuous map between connected spaces of the same dimension, which is a homeomorphism. In conclusion, a unique $\hat{p}$ minimizing $\mathbb{H}^{2}(r, p)$ subject to $\Xi(p)=\mathscr{M}_{2n}$ can be obtained by solving the dual problem.

\section{Statistical properties of the proposed density estimator}

In the previous sections, we proposed a novel parametrization of density function using power moments by the squared Hellinger distance. In this section, we analyze the statistical properties of the proposed estimator. By paraphrasing Theorem 4.5.5 in \cite{chung2001course}, we conclude the following theorem.

\begin{theorem} Denote the true density as $p$ and the corresponding random variable as $X$. Suppose there is a unique distribution function $F_{p}$ with the moments $\left\{\mu_{k}, k \geq\right.$ $1\}$, all finite. Denote the estimated density by power moments up to order $2n$ as $\hat{p}_{2n}$, and the corresponding random variable as $\hat{X}_{2n}$. Suppose that $\left(F_{{\hat{p}}_{2n}}\right)$ is a sequence of distribution functions, each of which has all its moments finite:
$$
\hat{\mu}_{2n, k}=\int_{-\infty}^{\infty} x^{k} d F_{\hat{p}_{2n}}.
$$

Then we have
$$
\mathbb{E}_{p}\left[\hat{\mu}_{2n,k}\right]=\mathbb{E}_{p}\left[\frac{1}{m} \sum_{j=1}^{m} X_{j}^{k}\right]=\frac{1}{m} \sum_{j=1}^{m} \mathbb{E}_{p}\left[X_{j}^{k}\right]=\mu_{k}.
$$

With $n \rightarrow +\infty$, the following equation holds for every $k \geq 1$ :
$$
\lim _{n \rightarrow \infty} \mathbb{E}_{p}\left[\hat{\mu}_{2n, k}\right]=\mu_{k}.
$$

Then $\hat{X}_{2n}$ converges to $X$ in distribution.
\label{theorem51}
\end{theorem}

Convergence in distribution is a relatively weak type of convergence, which requires the density estimate to be equal to the true density almost everywhere. Therefore Theorem \ref{theorem51} is indeed a weaker version of asymptotic unbiasedness, with $n \rightarrow +\infty$, where asymptotic unbiasedness is the convergence in probability. Here we emphasize that "asymptotic" refers to the number of moment terms used $2n \rightarrow +\infty$ rather than the number of samples $m \rightarrow +\infty$. Next we prove the consistency of the proposed estimator. Denote the estimation error as $\Delta p = \hat{p}_{2n} - p$ and write the Taylor expansion of it at $x = 0$ as
$$
    \Delta p = \sum_{k=0}^{+\infty} \frac{x^{k}}{k!}\Delta p^{(k)}(0).
$$

Then we write the estimation error in the L2 norm as
$$
\begin{aligned}
    & L_{2}\left( \hat{p}_{2n}, p \right)\\
    = & \int_{\mathbb{R}} \left( \Delta p \right)^{2} dx \\
    = & \int_{\mathbb{R}} \sum_{k=0}^{+\infty} \frac{x^{k}}{k!}\Delta p^{(k)}(0) \left( \hat{p}(x)- p(x) \right) dx \\
    = & \sum_{k=0}^{+\infty} \frac {\Delta p^{(k)}(0) }{k!} \int_{\mathbb{R}} x^{k} \left( \hat{p}(x)- p(x) \right) dx.
\end{aligned}
$$

As assumed in Theorem \ref{theorem51}, all power moments of both the density estimate and the true density exist and are finite. By denoting the $k_{\text{th}}$ order moment of $\hat{p}_{2n}, p_{2n}$ correspondingly as $\hat{\mu}_{k}, \mu_{k}, k \in \mathbb{N}_{0}$, we can write
$$
  L_{2}\left( \hat{p}_{2n}, p \right)
  = \sum_{k=0}^{+\infty} \frac{\Delta p^{(k)}(0) }{k!} \left( \hat{\mu}_{k} - \mu_{k} \right).
$$

By our proposed density surrogates, the first $2n+1$ power moments of $\hat{p}$ are identical to those of $p$, i.e. $\hat{\mu}_{k} = \mu_{k}$ for $k = 0, 1, \cdots, 2n$. Therefore we have
$$
    L_{2}\left( \hat{p}_{2n}, p \right) = \sum_{k=2n+1}^{+\infty} \frac{\Delta p^{(k)}(0)}{k!} \left( \hat{\mu}_{k} - \mu_{k} \right).
$$

Moreover, by the strong law of large numbers, we have
\begin{equation}
    \lim_{m \rightarrow \infty} \hat{\mu}_{k} = \lim_{m \rightarrow \infty} \frac{1}{m} \sum_{i = 1}^{m} X^{k} \stackrel{a.s.}{\longrightarrow} \mu_{k}, \quad k = 0, 1, 
    \cdots 2n.
\label{samplealmost}
\end{equation}

Therefore, we have
$$
    L_{2}\left( \hat{p}_{2n}, p \right) \stackrel{a.s.}{\longrightarrow}0,  \quad \text{with} \ n, m \rightarrow + \infty
$$
which shows that the proposed estimator is almost surely consistent in the sense of L2 norm \cite{izenman1991review, gordon1984almost}, given $n \rightarrow +\infty$.

\section{An asymptotic error upper bound of the estimator}

We will introduce an asymptotic upper bound of error for $\hat{p}(x)$, by the metric of total variation (TV) distance, which is a metric commonly adopted in the research on the moment problem.

The asymptotic TV distance reads
$$
\begin{aligned}
    & \lim_{m \rightarrow \infty}TV(\hat{p}, p)\\
    = & \lim_{m \rightarrow \infty} \sup_{x} \left|\int_{\left(-\infty, x \right] }(\hat{p} - p) d x\right|\\
    = & \lim_{m \rightarrow \infty} \sup_{x} \left| F_{\hat p} - F_{p} \right|
\end{aligned}
\label{lim}
$$
where $F_{\hat p}$ and $F_{p}$ are the two distribution functions of the density estimate $\hat{p}$ and the true density $p$.

Denote $\hat{p}_{t}$ as the density estimate using the true population moments of $p$, instead of the sample moments. Then by Theorem \ref{theorem51}, we have $
\lim_{m \rightarrow \infty}\hat{p} =  \hat{p}_{t}$ almost surely. Finally we have
$$
\lim_{m \rightarrow \infty}TV(\hat{p}, p)\stackrel{a.s.}{\longrightarrow} TV(\hat{p}_{t}, p).
$$

Shannon-entropy is adopted to derive an upper bound of the TV distance \cite{Aldo2003A}. It reads
$$
    H[p] = - \int_{\mathbb{R}}p(x) \log p(x)dx.
$$
We first introduce the Shannon-entropy maximizing distribution $F_{\breve{p}}$, of which the moments are the population moments of the true density. It has the following density function \cite{kapur1992entropy},
$$
    \breve{p}(x) = \exp \left ( - \sum_{i = 0}^{2n} m_{i} x^{i} \right )
$$
where $m_{0}, \cdots, m_{2n}$ are determined by the following constraints,
$$
    \int_{\mathbb{R}} x^k \exp \left ( - \sum_{i = 0}^{2n} m_{i} x^{i} \right )dx=\mu^{p}_{j}, \quad k=0,1, \cdots, 2n.
$$
By referring to \cite{Aldo2003A}, the KL distance between $p$ and $\breve{p}$ then reads
\begin{equation}
\begin{aligned}
    & KL \left(p\| \breve{p}\right)\\
    = & \int_{\mathcal{\mathbb{R}}} p(x) \log \frac{p(x)}{\breve{p}(x)} d x\\
    = & - H\left [ p \right ] + \sum_{i = 0}^{2n} m_{i} \mu^{p}_{j}\\
    = & H\left [ \breve{p} \right ] - H\left [ p \right ].
\end{aligned}
\label{KLBound1}
\end{equation}
The KL distance between $\hat{p}_{t}$ and $\breve{p}$ reads \begin{equation}
KL \left(\hat{p}_{t}\| \breve{p}\right) = H\left [ \breve{p} \right ] - H\left [ \hat{p}_{t} \right ].
\label{KLBound2}
\end{equation}

By \cite{1970Correction, Aldo2003A}, together with (\ref{KLBound1}), (\ref{KLBound2}) , we obtain
$$
\begin{aligned}
    & TV \left ( \breve{p}, \hat{p}_{t} \right )\\
    \leq & 3\left[\left(1+\frac{4}{9} KL \left(\hat{p}_{t} \| \breve{p} \right)\right)^{1 / 2}-1\right]^{1 / 2}\\
    = & 3\left[\left(1+\frac{4}{9} \left ( H\left [ \breve{p} \right ] - H\left [ \hat{p}_{t} \right ] \right )\right)^{1 / 2}-1\right]^{1 / 2}
\label{Vbound}
\end{aligned}
$$
and
$$
    TV \left ( \breve{p}, p \right ) \leq 3\left[\left(1+\frac{4}{9} \left ( H\left [ \breve{p} \right ] - H\left [ p \right ] \right )\right)^{1 / 2} - 1\right]^{1 / 2}.
$$
Then an asymptotic upper error bound  is given by
$$
\begin{aligned}
& TV \left ( \hat{p}_{t}, p \right )\\
= & \sup_{x}|F_{\hat{p}_{t}}\left ( x \right )-F_{p}\left ( x \right )|\\
\leq & \sup_{x}\left(\left|F_{\hat{p}_{t}}\left ( x \right )-F_{\breve{p}}\left ( x \right )\right|+\left|F_{\breve{p}}(x)-F_{p(x)}\right|\right)\\
\leq & \sup_{x} \left|F_{\hat{p}_{t}}\left ( x \right )-F_{\breve{p}}\left ( x \right )\right|+\sup_{x} \left|F_{\breve{p}}(x)-F_{p}(x)\right|\\
\leq & 3\left[\left(1+\frac{4}{9} \left ( H\left [ \breve{p} \right ] - H\left [ \hat{p}_{t} \right ] \right )\right)^{1 / 2}-1\right]^{1 / 2}\\
+ & 3\left[\left(1+\frac{4}{9} \left ( H\left [ \breve{p} \right ] - H\left [ p \right ] \right )\right)^{1 / 2} - 1\right]^{1 / 2}.
\end{aligned}
\label{UpperBoundUnbiased}
$$

If we are only given samples from the true density without knowing $p$, it is not possible for us to obtain the true $H[p]$. Under this circumstance, we approximate $H[p]$ by the empirical distribution function, which is $P\left( X = x_{i} \right) = r_{i}$. Then the Shannon entropy can be approximated as $H\left[ p \right] = - \sum r_{i} \log r_{i}$.

\section{Applications in signal processing tasks}

In the previous sections, we proposed an algorithm for density estimation using sample moments. However, the detailed mathematical treatments may have concealed the significance of the proposed density estimator. In \cite{kay2015probability}, a probability density estimator was proposed and applied to subset/feature selection. In this section, we will also introduce several applications of the proposed estimator in signal processing, and explain why they are indispensable in these applications. 

\subsection{Observation noise estimation for Bayesian filters}

In the conventional Kalman filter \cite{kalman1960new, kalman1961new} and its variants, including the extended Kalman filter (EKF) \cite{anderson2012optimal}, the central-difference Kalman filter (CDKF) \cite{schei1997finite, norgaard2000new}, the unscented Kalman filter (UKF) \cite{julier2000new}, and the quadrature Kalman filter (QKF) \cite{ito2000gaussian}, the probability density function of the additive observation noise is assumed to be Gaussian. Then estimating the density of the noise amounts to estimating the mean and variance of the noise. Given the samples of the noise, it is straightforward to obtain the sample mean and variance.

However, in real applications, the noises are always not Gaussian. In our recent papers \cite{wu2023non, wu2022multivariate}, we proposed a type of Bayesian filter based on power moments, where both the probability density functions of the system states and the observation noises can be non-Gaussian. Since the density function of the additive noise can be a non-Gaussian analytic one, the problem comes to estimating the observation noise given the samples from the noise. Since we are not provided with the model for estimation, conventional density estimation algorithms don't apply to this task. Furthermore, we desire the estimate to have a simple and analytic form of function for the ease of calculation, traditional nonparametric estimators are not proper options for this task either, due to the massive parameters. For example, given the i.i.d. samples $\left( X_{1}, X_{2}, \cdots, X_{m} \right)$ from the distribution of the observation noise, a typical kernel density estimator has the form

$$
\hat{p}(x) = \frac{1}{m}\sum_{i=1}^{m}K_{h}\left( x - X_{i} \right) = \frac{1}{mh}\sum_{i=1}^{m}K\left( \frac{x - X_{i}}{h} \right)
$$
where $K(\cdot)$ is the kernel function (a non-negative function), and $h > 0$ is a smoothing parameter called the bandwidth. Since we are always provided with hundreds even thousands of samples, the kernel density estimator needs to store the same amount of parameters, together with the parameters of the kernel function and the bandwidth. 

Compared with the kernel density estimators, our estimator has a much more compact form, where there are much less parameters to store. Assume that we use the first $2n$ orders of sample moments for density estimation. Our proposed estimator of the form \eqref{MiniForm} has only $2n$ parameters in $\hat{\Omega}$ and the parameters in the reference density $r(x)$. If we choose $r(x)$ to be a Gaussian density function, which is a pretty common choice, the number of parameters in the model is $2n+4$. The much fewer parameters greatly decreases the computation load for each filtering step, which reveals the significance of our proposed estimator.

\subsection{Recursive Bayesian estimator with non-Gaussian density functions}

We now consider a recursive Bayesian estimator with non-Gaussian density functions. Let the observation at time step $k$ be $u_{k}$. Denote the sequence of observations as 
$$
\mathbf{u}_{1: k} \triangleq\left\{u_i, i=1, \ldots, k\right\},
$$ 
with each $u_{i} \in \mathbb{R}$. The goal of the recursive Bayesian estimator is to estimate the posterior density function of the system state $y_{k}$, i.e., 
$$
p\left(y_{k} \mid \mathbf{u}_{1: k}\right).
$$

By Bayes formula, we have the following equation
\begin{equation}
\begin{aligned}
& p \left(y_{k} \mid \mathbf{u}_{1: k}\right)\\
= & \frac{1}{p\left(u_{k} \mid \mathbf{u}_{1: k-1}\right)} p\left(y_{k} \mid \mathbf{u}_{1: k-1}\right) p\left(u_{k} \mid y_{k}\right)\\
\propto & p\left(y_{k} \mid \mathbf{u}_{1: k-1}\right) p\left(u_{k} \mid y_{k}\right)
\label{inference}
\end{aligned}
\end{equation}
Now the problem amounts to calculating the right-hand-side of \eqref{inference} \cite{blom2007exact}. By the Chapman-Kolmogorov equation, we have
\begin{equation}
\begin{aligned}
& p\left(y_{k} \mid \mathbf{u}_{1: k-1}\right)\\
= & \int p\left(y_{k} \mid y_{k-1}\right) p\left(y_{k-1} \mid \mathbf{u}_{1: k-1}\right) d y_{k-1}.
\label{Prediction}
\end{aligned}
\end{equation}

We note that $p\left(y_{k} \mid y_{k-1}\right)$ and $p\left(u_{k} \mid y_{k}\right)$ can be directly determined by the system equation and the observation equation respectively. In the Bayesian estimation problem we treat, the probability densities are not necessarily Gaussian. Hence we are not always able to obtain an analytic $p\left(y_{k} \mid \mathbf{u}_{1: k-1}\right)$ by \eqref{Prediction}. We are then confronted with the problem of how to treat the possibly intractable integral \eqref{Prediction}. 

A common solution is to use the Monte-Carlo integration technique. The integral in \eqref{Prediction} can then be approximated by
\begin{equation}
\begin{aligned}
& p\left(y_{k} \mid \mathbf{u}_{1: k-1}\right)\\
\approx & \frac{\sum_{i=1}^{N} p\left(y_{k} \mid y_{k-1, i}\right) p\left(y_{k-1, i} \mid \mathbf{u}_{1: k-1}\right)}{\int \sum_{i=1}^{N} p\left(y_{k} \mid y_{k-1, i}\right) p\left(y_{k-1, i} \mid \mathbf{u}_{1: k-1}\right)dy_{k}}
\end{aligned}
\label{MonteCarlo}
\end{equation}
where $y_{k-1, i}$ for $i = 1, \cdots, N$ are $N$ i.i.d. samples uniformly drawn from its domain. By doing this, we note that $p\left(y_{k-1, i} \mid \mathbf{u}_{1: k-1}\right)$ is a probability value and the r.h.s. of \eqref{MonteCarlo} is now a weighted sum of $p\left(y_{k} \mid y_{k-1, i}\right)$. 

For a better estimation, $N$ needs to be selected as a large positive integer. However, it will cause the number of parameters in the estimated $p\left(y_{k} \mid \mathbf{u}_{1: k-1}\right)$ to be quite great, which makes the density estimate complicated. With the estimator proposed in the previous sections, we consider using the power moments to parameterize the density function to give a more compact representation of it. We note that it is not a difficult task to calculate the power moments of the density estimate, namely the r.h.s. of \eqref{MonteCarlo}. Then by properly selecting the number of power moments, we shall obtain a density estimate by power moments, which has the form \eqref{MiniForm}, with the algorithm proposed in the previous sections. 

In conclusion, density estimation is closely related to signal processing tasks, as proposed in this paper and in other previous ones. In the previous sections, we proposed the statistical properties and an asymptotic error upper bound of the density estimator. However, in the real applications of signal processing, people would be more interested in its real performance in different estimation tasks. As to better validate the performance of our proposed estimator, we perform three numerical simulations in the following section with a comparison to three prevailing estimators.

\section{Monte Carlo simulations}

This section reports the results of a Monte Carlo study designed to evaluate the performance of the proposed density estimator. We simulate mixtures of probability density functions, including Gaussian and non-Gaussian, smooth and non-smooth. These simulations validate the ability of the proposed density estimator as applied to much wider classes of functions. 

We give performance comparisons of the following algorithms. First is the estimate by the density parametrization using moments by squared Hellinger distance (DPMSH), of which the curves are colored blue in all the subsequent figures. The orange curves are those of estimates by the density parametrization using moments by Kullback-Leibler distance (DPMKL), which was proposed in \cite{wu2023non}. The green curves represent the estimates by a typical kernel density estimator (KDE), of which the kernel function is chosen as Gaussian and the corresponding bandwidth is chosen by Silverman's bandwidth selection. The red curves are the ones by the Gaussian mixture model (GMM) where the number of modes is set to be two for the five examples. We note that since previous methods of moments are not able to treat the density estimation problem without knowledge of the number of modes or feasible function class, we don't compare them to our proposed algorithm in this paper. In each of the following examples, a figure showing the true density and the average density estimates by the four algorithms, a figure showing the average TV distances between the true densities and the estimates over different number of samples, and one showing the Kullback-Leibler distance between the true density and the estimates shall be given for a complete comparison between the four algorithms.

The reference density $r$ can usually be chosen as a Gaussian $r(x)=\mathcal{N}\left(m, \sigma^{2}\right)$, of which the parameters can be selected as $m=\mu_{1}$ and $\sigma^{2}>\mu_{2}$. $\mu_{1}, \mu_{2}$ can be calculated by \eqref{MomentCal}. Here we note that a relatively large variance $\sigma^{2}$ always yields better estimation performance for the density functions which have multiple peaks (modes).

The first example is a mixture of two Gaussians
$$
     p(x) = \frac{0.5}{\sqrt{2\pi}}\exp \left({\frac{(x-2)^{2}}{2}} \right) +  \frac{0.5}{\sqrt{2\pi}}\exp \left({\frac{(x+2)^{2}}{2}} \right).
$$

The prior $r$ is chosen as a Gaussian distribution $\mathcal{N}(0, 6.7^{2})$. The simulation results are given in Figure 1-3. Figure 1 shows the average density estimates of $50$ Monte Carlo simulations with $100$ data samples, i.e. $\mathbb{E}_{p}[\hat{p}(x)]$, which is used in density estimation to show the unbiasedness \cite{izenman1991review}. Figure 2 shows the TV distances between the density estimates and the true density by the four methods with different number of data samples. Figure 3 shows the Kullback-Leibler distances with different number of data samples. We observe in the left image that the average estimate by GMM is closest to the true density. However it is partly due to the prior knowledge that there are two Gaussians in the true density. We also note that the estimates by KDE suffer from the lack of data samples. The density estimate by DPMSH in this example uses the sample moments up to order $4$. It has the second best performance, in the senses of both the TV distance and the Kullback-Leibler distance. We emphasize that unlike GMM, our proposed density estimator doesn't have prior knowledge of the true density to be estimated, e.g. the number of modes or the feasible function classes. As we mentioned in the previous sections, DPMKL has sharp peaks due to using the Kullback-Leibler distance.

\begin{figure}[htbp]
\centering
\includegraphics[scale=0.38]{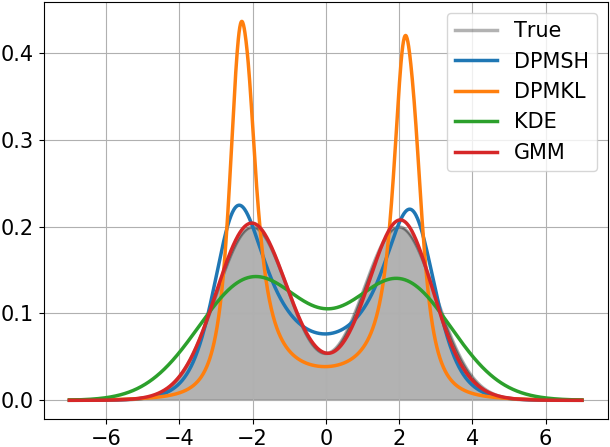}
\centering
\caption{Simulation results of Example 1. Average density estimates of $50$ Monte Carlo simulations with $100$ data samples. }
\label{fig11}
\end{figure}

\begin{figure}[htbp]
\centering
\includegraphics[scale=0.38]{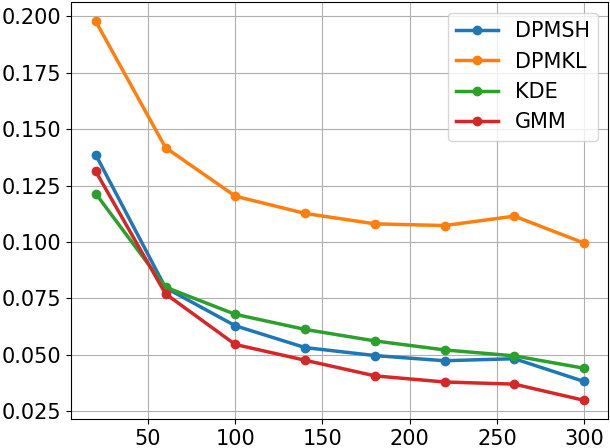}
\centering
\caption{Simulation results of Example 1. The total variation distances between the density estimates and the true density over different number of samples.}
\label{fig12}
\end{figure}

\begin{figure}[htbp]
\centering
\includegraphics[scale=0.38]{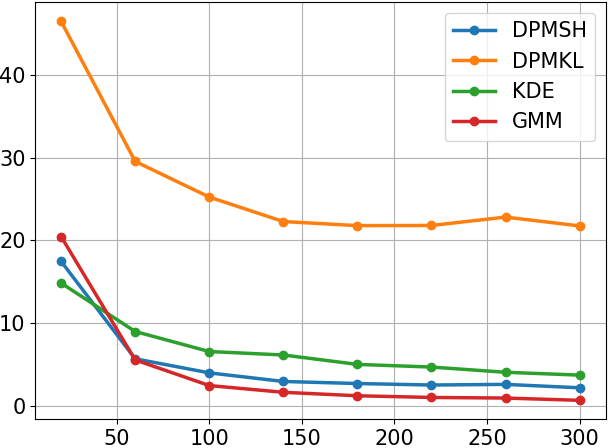}
\centering
\caption{Simulation results of Example 1. The Kullback-Leibler distances between the density estimates and the true density over different number of samples.}
\label{fig13}
\end{figure}

The second example is another mixture of Gaussians
$$
     p(x) = \frac{0.7}{\sqrt{2\pi}}\exp \left({\frac{(x-2)^{2}}{2}} \right) +  \frac{0.3}{\sqrt{2\pi}}\exp \left({\frac{(x+2)^{2}}{2}} \right).
$$

We design this example to test the ability of the proposed estimator in estimating modes with small values of probability. The prior $r$ is chosen as a Gaussian distribution $\mathcal{N}(-0.7, 6.2^{2})$. The simulation results are given in Figure 4-6. Figure 4 shows the average density estimates of $50$ Monte Carlo simulations with $100$ data samples. GMM has the best performance. KDE and DPMSH have comparable performances in the senses of both the TV distance and the KL distance. KDE model stores the same number of the parameters as the data samples. However there are only $5$ parameters in our proposed DPMSH model, where $2n=4$ in this example. It reveals the advantage of our proposed DPMSH over other methods.

\begin{figure}[htbp]
\centering
\includegraphics[scale=0.38]{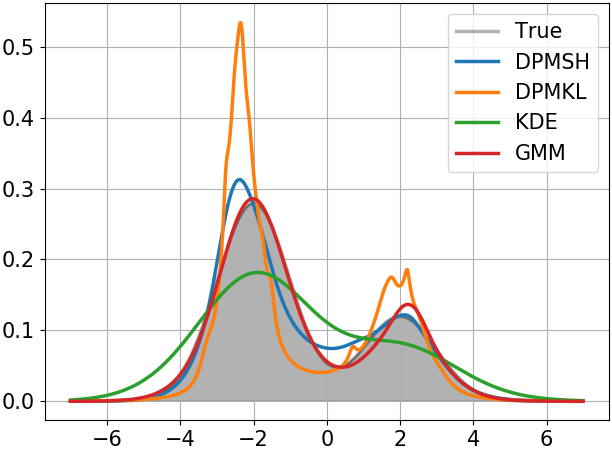}
\centering
\caption{Simulation results of Example 2. Average density estimates of $50$ Monte Carlo simulations with $100$ data samples. }
\label{fig21}
\end{figure}

\begin{figure}[htbp]
\centering
\includegraphics[scale=0.38]{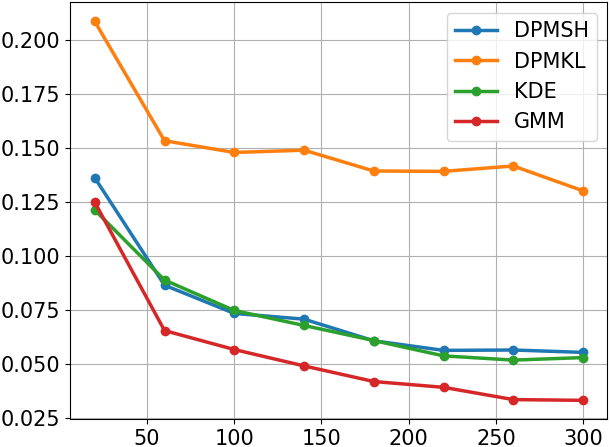}
\centering
\caption{Simulation results of Example 2. The TV distances between the estimators and the true density.}
\label{fig22}
\end{figure}

\begin{figure}[htbp]
\centering
\includegraphics[scale=0.38]{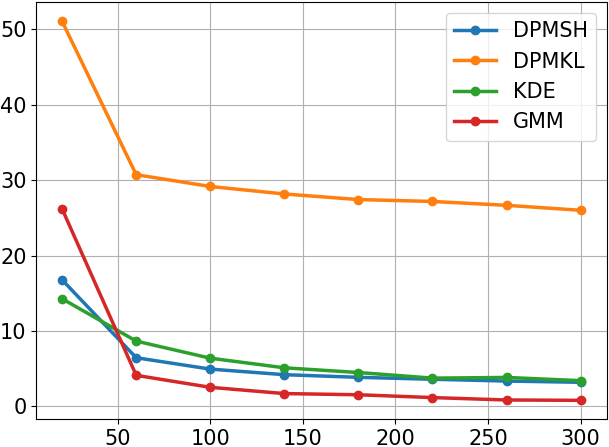}
\centering
\caption{Simulation results of Example 2. The KL distances between the estimators and the true density.}
\label{fig23}
\end{figure}

In the following two examples, we simulate on mixtures of non-Gaussian densities. Example 3 simulates a mixture of two Laplace distributions
\begin{equation*}
     p(x) = 0.5 \exp \left({-2\left| x - 2\right|}\right) + 0.5 \exp \left({-2\left| x + 2 \right|}\right).    
\end{equation*}

The prior $r$ is chosen as a Gaussian distribution $\mathcal{N}(0, 6.5^{2})$. The simulation results are given in Figure 7-9. Figure 7 shows the average density estimate of $50$ Monte Carlo simulations with $200$ data samples. We note that the performance of the density estimate by DPMSH using sample moments up to order $4$ is better than KDE without prior knowledge of the number of modes.

\begin{figure}[htbp]
\centering
\includegraphics[scale=0.38]{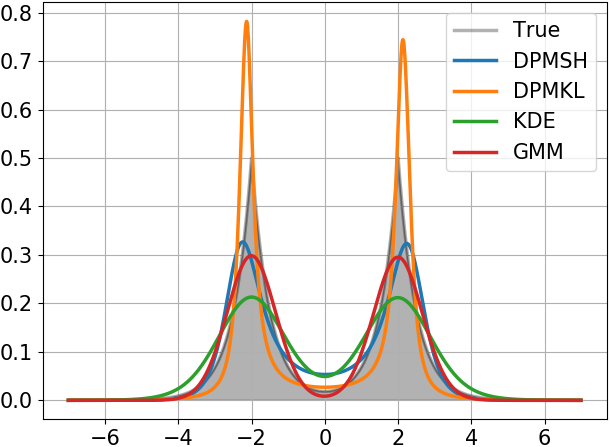}
\centering
\caption{Simulation results of Example 3. Average density estimates of $50$ Monte Carlo simulations with $200$ data samples.}
\label{fig31}
\end{figure}

\begin{figure}[htbp]
\centering
\includegraphics[scale=0.38]{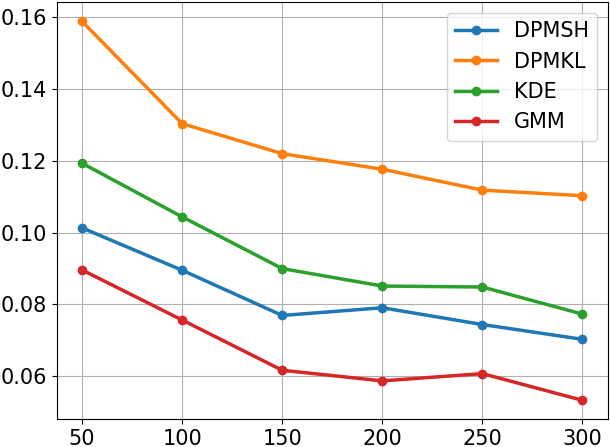}
\centering
\caption{Simulation result of Example 3. The TV distances between the estimators and the true density.}
\label{fig32}
\end{figure}

\begin{figure}[htbp]
\centering
\includegraphics[scale=0.38]{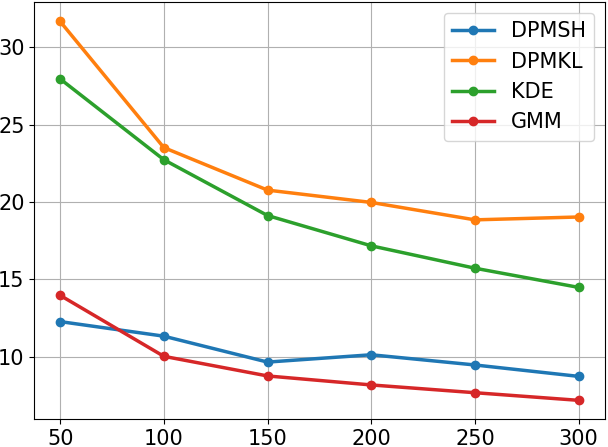}
\centering
\caption{Simulation result of Example 3. The KL distances between the estimators and the true density.}
\label{fig33}
\end{figure}

Example 4 is a mixture of two Gumbel distributions
$$
\begin{aligned}
     p(x) = & 0.5\exp\left(-\left(x - 1 + \exp\left(- (x - 1) \right)\right)\right)\\
     + & 0.5\exp\left(-\left(x + 1 + \exp\left(- (x + 1) \right)\right)\right)
\end{aligned}
$$

The prior $r$ is chosen as a Gaussian distribution $\mathcal{N}(0.5, 3.5^{2})$. The simulation results are given in Figure 10-12, which are the average of $50$ Monte Carlo simulations with $200$ data samples. In this example, the two modes are not easy to distinguish. Our proposed DPMSH, which uses sample moments up to order $6$, obtains the best performance comparable to KDE. Since in this example, the prior constraint of the densities being Gaussian is no longer valid for GMM, the estimation performance of it is not as good as that of DPMSH. Moreover, except for the DPMKL estimate which has two distinct modes but is not close to the true density, only DPMSH approximates the two modes in the remaining three methods. 

\begin{figure}[htbp]
\centering
\includegraphics[scale=0.38]{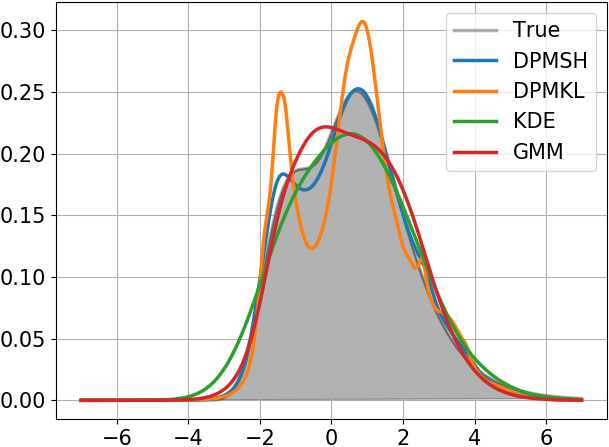}
\centering
\caption{Simulation result of Example 4. Average density estimates of $50$ Monte Carlo simulations with $200$ data samples.}
\label{fig1}
\end{figure}

\begin{figure}[htbp]
\centering
\includegraphics[scale=0.38]{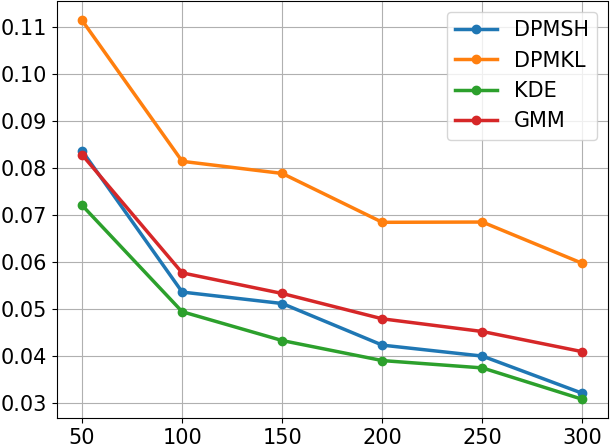}
\centering
\caption{Simulation result of Example 4. The TV distances between the estimators and the true density.}
\label{fig2}
\end{figure}

\begin{figure}[htbp]
\centering
\includegraphics[scale=0.38]{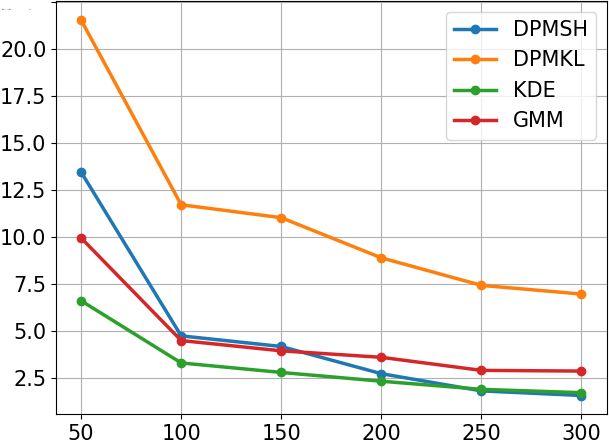}
\centering
\caption{Simulation result of Example 4. The KL distances between the estimators and the true density.}
\label{fig3}
\end{figure}

Last we simulate the case where the number of densities in the mixture is larger than the number of modes. Example 5 is a mixture of 3 Gaussians, however there are only 2 modes,
$$
\begin{aligned}
     p(x) = & \frac{0.3}{\sqrt{2\pi}}\exp \left({\frac{(x-3)^{2}}{2}} \right) +  \frac{0.3}{\sqrt{2\pi}}\exp \left({\frac{(x+3)^{2}}{2}} \right)\\
     + & \frac{0.4}{\sqrt{2\pi} \cdot 2}\exp \left({\frac{(x-1)^{2}}{2 \cdot 4}} \right).
\end{aligned}
$$

The prior $r$ is chosen as a Gaussian distribution $\mathcal{N}(0.3, 5.0^{2})$. The simulation results are given in Figure 13-15, which are the average of $50$ Monte Carlo simulations with $200$ data samples. In this example, we use sample moments up to order $6$. We note that the performance of our proposed DPMSH estimate achieves the best performance. This example reveals the ability of our proposed parameterizaiton in estimating the modes which are a mixture of densities.

\begin{figure}[htbp]
\centering
\includegraphics[scale=0.38]{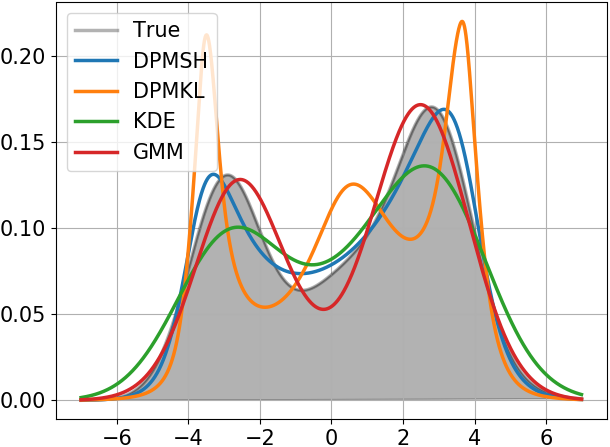}
\centering
\caption{Simulation result of Example 5. Average density estimates of $50$ Monte Carlo simulations with $200$ data samples.}
\label{fig51}
\end{figure}

\begin{figure}[htbp]
\centering
\includegraphics[scale=0.38]{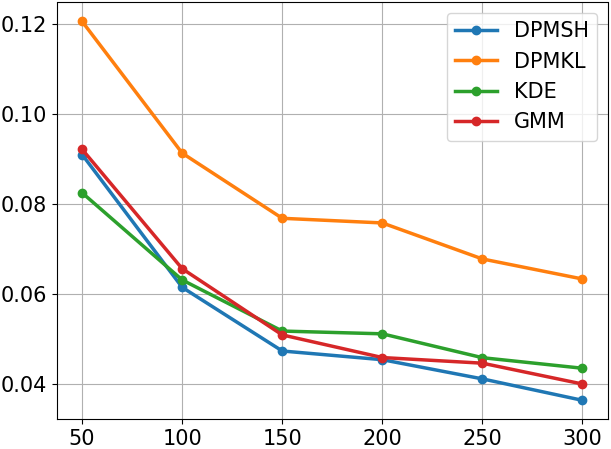}
\centering
\caption{Simulation result of Example 5. The TV distances between the estimators and the true density.}
\label{fig52}
\end{figure}

\begin{figure}[htbp]
\centering
\includegraphics[scale=0.38]{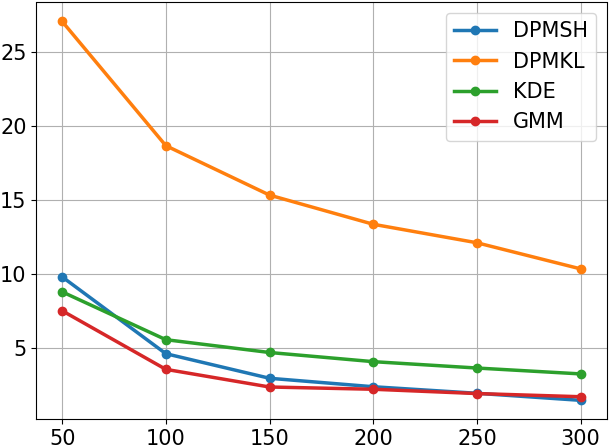}
\centering
\caption{Simulation result of Example 5. The KL distances between the estimators and the true density.}
\label{fig53}
\end{figure}

\section{Conclusion}

We have developed an algorithm to parameterize and estimate probability density $p(x)$ on the real line from sample power moments by the squared Hellinger distance, leading to feasible solutions of the form (\ref{MiniForm}). No prior constraints are imposed on the density to be estimated, such as a prescribed mixture of densities.  The parametrization is in terms of a general prior density $r(x)$ with no particular connection to the data, generally chosen to be Gaussian. For each choice of prior $r(x)$ we obtain an analytic form the density estimate which is closest to $r(x)$ in the squared Hellinger distance. The map $\zeta: \mathcal{R}_{+} \rightarrow \mathcal{X}_{+}$ is proved to be homeomorphic, which establishes the existence and uniqueness of the solution. This also provides a convex optimization problem with the cost functional (\ref{CostFunc}). Moreover, we propose statistical properties and an asymptotic error upper bound for the density estimator using power moments. Two of its important applications in the signal processing tasks are given, together with an explanation of its indispensability in these tasks. The simulation results on multi-modal density estimation also show the performance of the proposed estimator without prior information or estimation of the number of modes or the feasible classes of the density. The theoretical proofs and the simulation results both reveal the significance of the non-classical parametrization by power moments.

\bibliographystyle{IEEEtran}
\bibliography{references}

% Generated by IEEEtran.bst, version: 1.14 (2015/08/26)
\begin{thebibliography}{10}
\providecommand{\url}[1]{#1}
\csname url@samestyle\endcsname
\providecommand{\newblock}{\relax}
\providecommand{\bibinfo}[2]{#2}
\providecommand{\BIBentrySTDinterwordspacing}{\spaceskip=0pt\relax}
\providecommand{\BIBentryALTinterwordstretchfactor}{4}
\providecommand{\BIBentryALTinterwordspacing}{\spaceskip=\fontdimen2\font plus
\BIBentryALTinterwordstretchfactor\fontdimen3\font minus
  \fontdimen4\font\relax}
\providecommand{\BIBforeignlanguage}[2]{{%
\expandafter\ifx\csname l@#1\endcsname\relax
\typeout{** WARNING: IEEEtran.bst: No hyphenation pattern has been}%
\typeout{** loaded for the language `#1'. Using the pattern for}%
\typeout{** the default language instead.}%
\else
\language=\csname l@#1\endcsname
\fi
#2}}
\providecommand{\BIBdecl}{\relax}
\BIBdecl

\bibitem{parzen1962estimation}
E.~Parzen, ``On estimation of a probability density function and mode,''
  \emph{The annals of mathematical statistics}, vol.~33, no.~3, pp. 1065--1076,
  1962.

\bibitem{silverman2018density}
B.~W. Silverman, \emph{Density estimation for statistics and data
  analysis}.\hskip 1em plus 0.5em minus 0.4em\relax Routledge, 2018.

\bibitem{mclachlan1988mixture}
G.~J. McLachlan and K.~E. Basford, \emph{Mixture models: Inference and
  applications to clustering}.\hskip 1em plus 0.5em minus 0.4em\relax M. Dekker
  New York, 1988, vol.~38.

\bibitem{bunea2007sparse}
F.~Bunea, A.~B. Tsybakov, and M.~H. Wegkamp, ``Sparse density estimation with
  $l$1 penalties,'' in \emph{International Conference on Computational Learning
  Theory}.\hskip 1em plus 0.5em minus 0.4em\relax Springer, 2007, pp. 530--543.

\bibitem{barndorff2014information}
O.~Barndorff-Nielsen, \emph{Information and exponential families: in
  statistical theory}.\hskip 1em plus 0.5em minus 0.4em\relax John Wiley \&
  Sons, 2014.

\bibitem{dudik2004performance}
M.~Dudik, S.~J. Phillips, and R.~E. Schapire, ``Performance guarantees for
  regularized maximum entropy density estimation,'' in \emph{International
  Conference on Computational Learning Theory}.\hskip 1em plus 0.5em minus
  0.4em\relax Springer, 2004, pp. 472--486.

\bibitem{altun2006unifying}
Y.~Altun and A.~Smola, ``Unifying divergence minimization and statistical
  inference via convex duality,'' in \emph{International Conference on
  Computational Learning Theory}.\hskip 1em plus 0.5em minus 0.4em\relax
  Springer, 2006, pp. 139--153.

\bibitem{song2008tailoring}
L.~Song, X.~Zhang, A.~Smola, A.~Gretton, and B.~Sch{\"o}lkopf, ``Tailoring
  density estimation via reproducing kernel moment matching,'' in
  \emph{Proceedings of the 25th international conference on Machine learning},
  2008, pp. 992--999.

\bibitem{chernoff1964estimation}
H.~Chernoff, ``Estimation of the mode,'' \emph{Annals of the Institute of
  Statistical Mathematics}, vol.~16, no.~1, pp. 31--41, 1964.

\bibitem{eddy1980optimum}
W.~F. Eddy, ``Optimum kernel estimators of the mode,'' \emph{The Annals of
  Statistics}, vol.~8, no.~4, pp. 870--882, 1980.

\bibitem{cheng1995mean}
Y.~Cheng, ``Mean shift, mode seeking, and clustering,'' \emph{IEEE transactions
  on pattern analysis and machine intelligence}, vol.~17, no.~8, pp. 790--799,
  1995.

\bibitem{abraham2004asymptotic}
C.~Abraham, G.~Biau, and B.~Cadre, ``On the asymptotic properties of a simple
  estimate of the mode,'' \emph{ESAIM: Probability and Statistics}, vol.~8, pp.
  1--11, 2004.

\bibitem{dasgupta2014optimal}
S.~Dasgupta and S.~Kpotufe, ``Optimal rates for k-nn density and mode
  estimation,'' \emph{Advances in Neural Information Processing Systems},
  vol.~27, pp. 2555--2563, 2014.

\bibitem{genovese2012minimax}
C.~R. Genovese, M.~P. Pacifico, I.~Verdinelli, L.~Wasserman \emph{et~al.},
  ``Minimax manifold estimation,'' \emph{Journal of machine learning research},
  vol.~13, pp. 1263--1291, 2012.

\bibitem{jiang2017modal}
H.~Jiang and S.~Kpotufe, ``Modal-set estimation with an application to
  clustering,'' in \emph{Artificial Intelligence and Statistics}.\hskip 1em
  plus 0.5em minus 0.4em\relax PMLR, 2017, pp. 1197--1206.

\bibitem{rigollet2007generalization}
P.~Rigollet, ``Generalization error bounds in semi-supervised classification
  under the cluster assumption.'' \emph{Journal of Machine Learning Research},
  vol.~8, no.~7, 2007.

\bibitem{georgiou2003kullback}
T.~T. Georgiou and A.~Lindquist, ``Kullback-{L}eibler approximation of spectral
  density functions,'' \emph{IEEE Transactions on Information Theory}, vol.~49,
  no.~11, pp. 2910--2917, 2003.

\bibitem{schmudgen2017moment}
K.~Schm{\"u}dgen, \emph{The moment problem}.\hskip 1em plus 0.5em minus
  0.4em\relax Graduate Texts in Mathematics, 2017, vol. 277.

\bibitem{bertsimas2005optimal}
D.~Bertsimas and I.~Popescu, ``Optimal inequalities in probability theory: A
  convex optimization approach,'' \emph{SIAM Journal on Optimization}, vol.~15,
  no.~3, pp. 780--804, 2005.

\bibitem{wu2023non}
G.~Wu and A.~Lindquist, ``Non-{G}aussian {B}ayesian filtering by density
  parametrization using power moments,'' \emph{Automatica}, vol. 153, p.
  111061, 2023.

\bibitem{hall1987kullback}
P.~Hall, ``On kullback-{L}eibler loss and density estimation,'' \emph{The
  Annals of Statistics}, pp. 1491--1519, 1987.

\bibitem{li1999mixture}
J.~Q. Li and A.~R. Barron, ``Mixture {D}ensity {E}stimation.'' in \emph{NIPS},
  vol.~12, 1999, pp. 279--285.

\bibitem{vapnik1999nature}
V.~Vapnik, \emph{The nature of statistical learning theory}.\hskip 1em plus
  0.5em minus 0.4em\relax Springer science \& business media, 1999.

\bibitem{cutler1996minimum}
A.~Cutler and O.~I. Cordero-Brana, ``Minimum {H}ellinger distance estimation
  for finite mixture models,'' \emph{Journal of the American Statistical
  association}, vol.~91, no. 436, pp. 1716--1723, 1996.

\bibitem{lu2003minimum}
Z.~Lu, Y.~V. Hui, and A.~H. Lee, ``Minimum {H}ellinger distance estimation for
  finite mixtures of poisson regression models and its applications,''
  \emph{Biometrics}, vol.~59, no.~4, pp. 1016--1026, 2003.

\bibitem{chung2001course}
K.~L. Chung, \emph{A course in probability theory}.\hskip 1em plus 0.5em minus
  0.4em\relax Academic press, 2001.

\bibitem{izenman1991review}
A.~J. Izenman, ``Review papers: Recent developments in nonparametric density
  estimation,'' \emph{Journal of the american statistical association},
  vol.~86, no. 413, pp. 205--224, 1991.

\bibitem{gordon1984almost}
L.~Gordon and R.~A. Olshen, ``Almost surely consistent nonparametric regression
  from recursive partitioning schemes,'' \emph{Journal of Multivariate
  Analysis}, vol.~15, no.~2, pp. 147--163, 1984.

\bibitem{Aldo2003A}
A.~Tagliani, ``A note on proximity of distributions in terms of coinciding
  moments,'' \emph{Applied Mathematics and Computation}, vol. 145, no. 2-3, pp.
  195--203, 2003.

\bibitem{kapur1992entropy}
J.~N. Kapur and H.~K. Kesavan, ``Entropy optimization principles and their
  applications,'' in \emph{Entropy and energy dissipation in water
  resources}.\hskip 1em plus 0.5em minus 0.4em\relax Springer, 1992, pp. 3--20.

\bibitem{1970Correction}
S.~Kullback, ``Correction to a lower bound for discrimination information in
  terms of variation,'' \emph{IEEE Transactions on Information Theory},
  vol.~16, no.~5, pp. 652--652, 1970.

\bibitem{kay2015probability}
S.~Kay, Q.~Ding, B.~Tang, and H.~He, ``Probability density function estimation
  using the eef with application to subset/feature selection,'' \emph{IEEE
  Transactions on Signal Processing}, vol.~64, no.~3, pp. 641--651, 2015.

\bibitem{kalman1960new}
R.~E. Kalman, ``A new approach to linear filtering and prediction problems,''
  \emph{Journal of Basic Engineering}, vol.~82, no.~1, pp. 35--45, 1960.

\bibitem{kalman1961new}
R.~E. Kalman and R.~S. Bucy, ``New results in linear filtering and prediction
  theory,'' \emph{Journal of Basic Engineering}, vol.~83, no.~1, pp. 95--108,
  1961.

\bibitem{anderson2012optimal}
B.~D. Anderson and J.~B. Moore, \emph{Optimal filtering}.\hskip 1em plus 0.5em
  minus 0.4em\relax Courier Corporation, 2012.

\bibitem{schei1997finite}
T.~S. Schei, ``A finite-difference method for linearization in nonlinear
  estimation algorithms,'' \emph{Automatica}, vol.~33, no.~11, pp. 2053--2058,
  1997.

\bibitem{norgaard2000new}
M.~Norgaard, N.~K. Poulsen, and O.~Ravn, ``New developments in state estimation
  for nonlinear systems,'' \emph{Automatica}, vol.~36, no.~11, pp. 1627--1638,
  2000.

\bibitem{julier2000new}
S.~Julier, J.~Uhlmann, and H.~F. Durrant-Whyte, ``A new method for the
  nonlinear transformation of means and covariances in filters and
  estimators,'' \emph{IEEE Transactions on automatic control}, vol.~45, no.~3,
  pp. 477--482, 2000.

\bibitem{ito2000gaussian}
K.~Ito and K.~Xiong, ``Gaussian filters for nonlinear filtering problems,''
  \emph{IEEE transactions on automatic control}, vol.~45, no.~5, pp. 910--927,
  2000.

\bibitem{wu2022multivariate}
G.~Wu and A.~Lindquist, ``A multivariate non-{G}aussian {B}ayesian filter using
  power moments,'' \emph{arXiv preprint arXiv:2211.13374}, 2022.

\bibitem{blom2007exact}
H.~A. Blom and E.~A. Bloem, ``Exact bayesian and particle filtering of
  stochastic hybrid systems,'' \emph{IEEE Transactions on Aerospace and
  Electronic Systems}, vol.~43, no.~1, pp. 55--70, 2007.

\end{thebibliography}

\newpage

\begin{IEEEbiography}[{\includegraphics[width=1in,height=1.25in,clip,keepaspectratio]{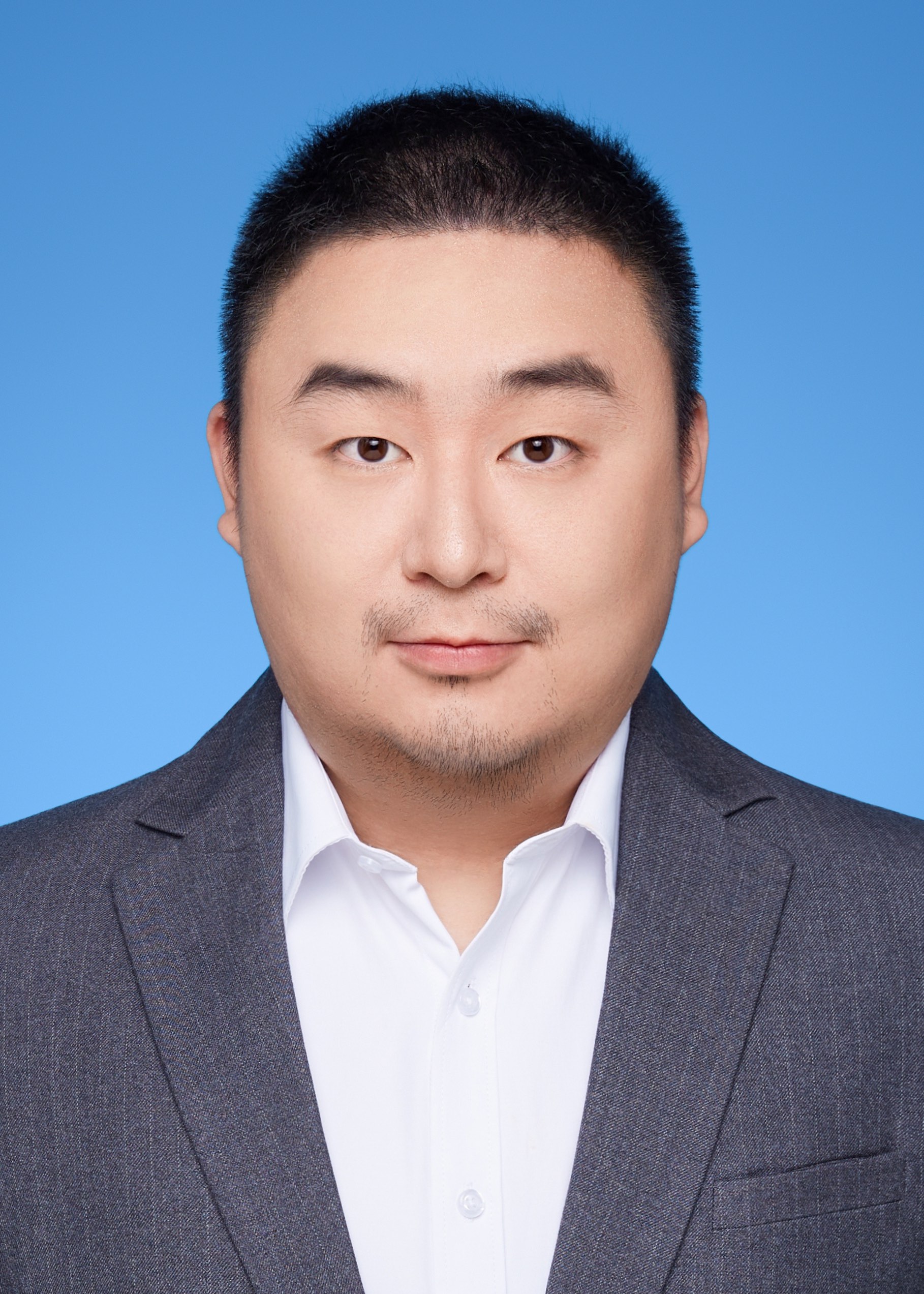}}]{Guangyu Wu} (S'22) received the B.E. degree from Northwestern Polytechnical University, Xi’an, China, in 2013, and two M.S. degrees, one in control science and engineering from Shanghai Jiao Tong University, Shanghai, China, in 2016, and the other in electrical engineering from the University of Notre Dame, South Bend, USA, in 2018. He is currently pursuing the Ph.D. degree at Shanghai Jiao Tong University.

\end{IEEEbiography}

\begin{IEEEbiography}[{\includegraphics[width=1in,height=1.25in,clip,keepaspectratio]{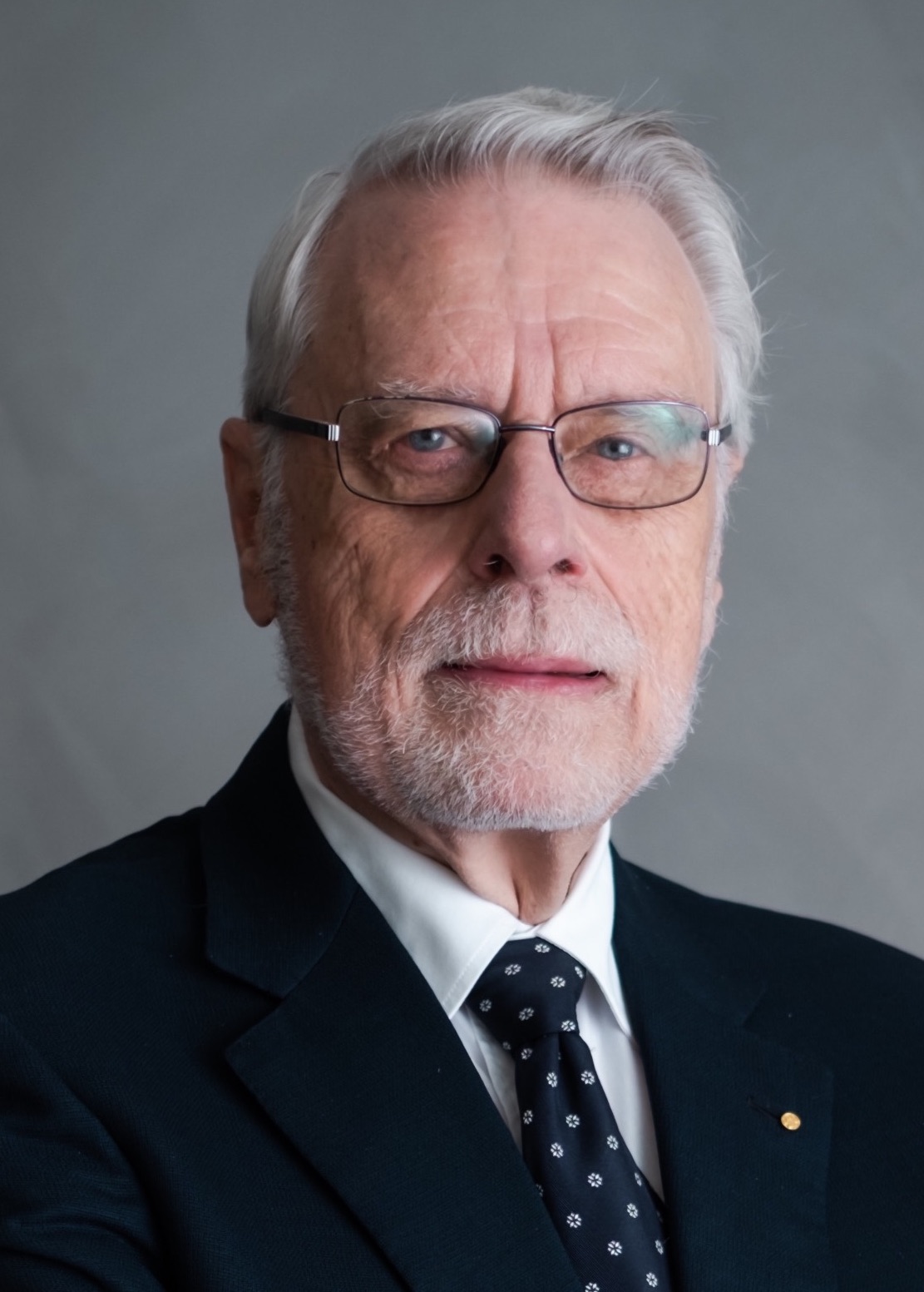}}]{Anders Lindquist} (M’77–SM’86–F’89–LF’10) received the Ph.D. degree in optimization and systems theory from the Royal Institute of Technology (KTH), Stockholm, Sweden, in 1972, an honorary doctorate (Doctor Scientiarum Honoris Causa) from Technion (Israel Institute of Technology) in 2010, and Doctor Jubilaris from KTH in 2022. 

He is currently a Zhiyuan Chair Professor at Shanghai Jiao Tong University, China, and Professor Emeritus at KTH, Stockholm, Sweden. Before that he had a full academic career in the United States, after which he was appointed to the Chair of Optimization and Systems at KTH.

Dr. Lindquist is a Member of the Royal Swedish Academy of Engineering Sciences, a Foreign Member of the Chinese Academy of Sciences, a Member of Academia Europaea (Academy of Europe), an Honorary Member the Hungarian Operations Research Society, a Life Fellow of IEEE, a Fellow of SIAM, and a Fellow of IFAC. He received the 2003 George S. Axelby Outstanding Paper Award, the 2009 Reid Prize in Mathematics from SIAM, and the 2020 IEEE Control Systems Award, the IEEE field award in Systems and Control.
\end{IEEEbiography}

\vfill

\end{document}